\documentclass{article}
\usepackage[final]{neurips_2024}

\input{config.sty}
\input{glossary.sty}

\title{Expert-level protocol translation for self-driving labs}

\author{
Yu-Zhe Shi$^{\star{}}$, Fanxu Meng$^{\star{}}$, Haofei Hou$^{\star{}}$, Zhangqian Bi,  Qiao Xu,\\ 
\textbf{Lecheng Ruan}$^{\text{\Letter}}$, \textbf{Qining Wang}$^{\text{\Letter}}$
\vspace{0.5em}\\
Department of Advanced Manufacturing and Robotics, \\College of Engineering, Peking University \quad{}\\
$^\star{}$Equal contribution\quad
\Letter\phantom\,\,\texttt{ruanlecheng@ucla.edu, qiningwang@pku.edu.cn}
\vspace{-9pt}
}

\begin{document}

\maketitle

\begin{abstract}
Recent development in \ac{ai} models has propelled their application in scientific discovery, but the validation and exploration of these discoveries require subsequent empirical experimentation. The concept of self-driving laboratories promises to automate and thus boost the experimental process following \ac{ai}-driven discoveries. However, the transition of experimental protocols, originally crafted for human comprehension, into formats interpretable by machines presents significant challenges, which, within the context of specific expert domain, encompass the necessity for structured as opposed to natural language, the imperative for explicit rather than tacit knowledge, and the preservation of causality and consistency throughout protocol steps. Presently, the task of protocol translation predominantly requires the manual and labor-intensive involvement of domain experts and information technology specialists, rendering the process time-intensive. To address these issues, we propose a framework that automates the protocol translation process through a three-stage workflow, which incrementally constructs \acp{pdg} that approach structured on the syntax level, completed on the semantics level, and linked on the execution level. Quantitative and qualitative evaluations have demonstrated its performance at par with that of human experts, underscoring its potential to significantly expedite and democratize the process of scientific discovery by elevating the automation capabilities within self-driving laboratories.
\end{abstract}

\section{Introduction}

The evolution of \ac{ai} techniques has significantly accelerated the processes inherent to scientific discovery, with a notable impact observed within the domain of experimental sciences~\citep{wang2023scientific}. This influence is manifested through a variety of avenues: the generation of hypothesis spaces informed by extensive literature analysis~\citep{jablonka2022making,kim2024transparent}, the interpretation of observational data via the identification of high-dimensional correlations~\citep{jumper2021highly,abramson2024accurate}, the engineering of novel structures that meet predefined specifications~\citep{grisoni2021combining,park2023artificial}, and the implementation of comprehensive simulations to ascertain the characteristics of potential products~\citep{hie2021learning,singh2023emergent}. 

However, the findings facilitated by AI-driven research require further validation and exploration via empirical experiments, and may even entail a cyclical process where AI-generated hypotheses are refined based on the outcomes of real-world experiments, which demands the assembly of a sizable cohort of experienced experimenters to carry out these investigations in accordance with established \emph{protocols}~\citep{mcnutt2014reproducibility}. Unfortunately, the formation and sustenance of such a dedicated experimental cadre are fraught with considerable financial demands, and the collaborative engagement between people oriented towards AI methodologies and those grounded in experimental sciences is frequently encumbered by the communication gaps between distinct intellectual paradigms~\citep{baker20161,freedman2015economics,munafo2017manifesto,baker2021five,shi2023perslearn}.

\begin{figure}[t]
    \centering
    \includegraphics[width=\textwidth]{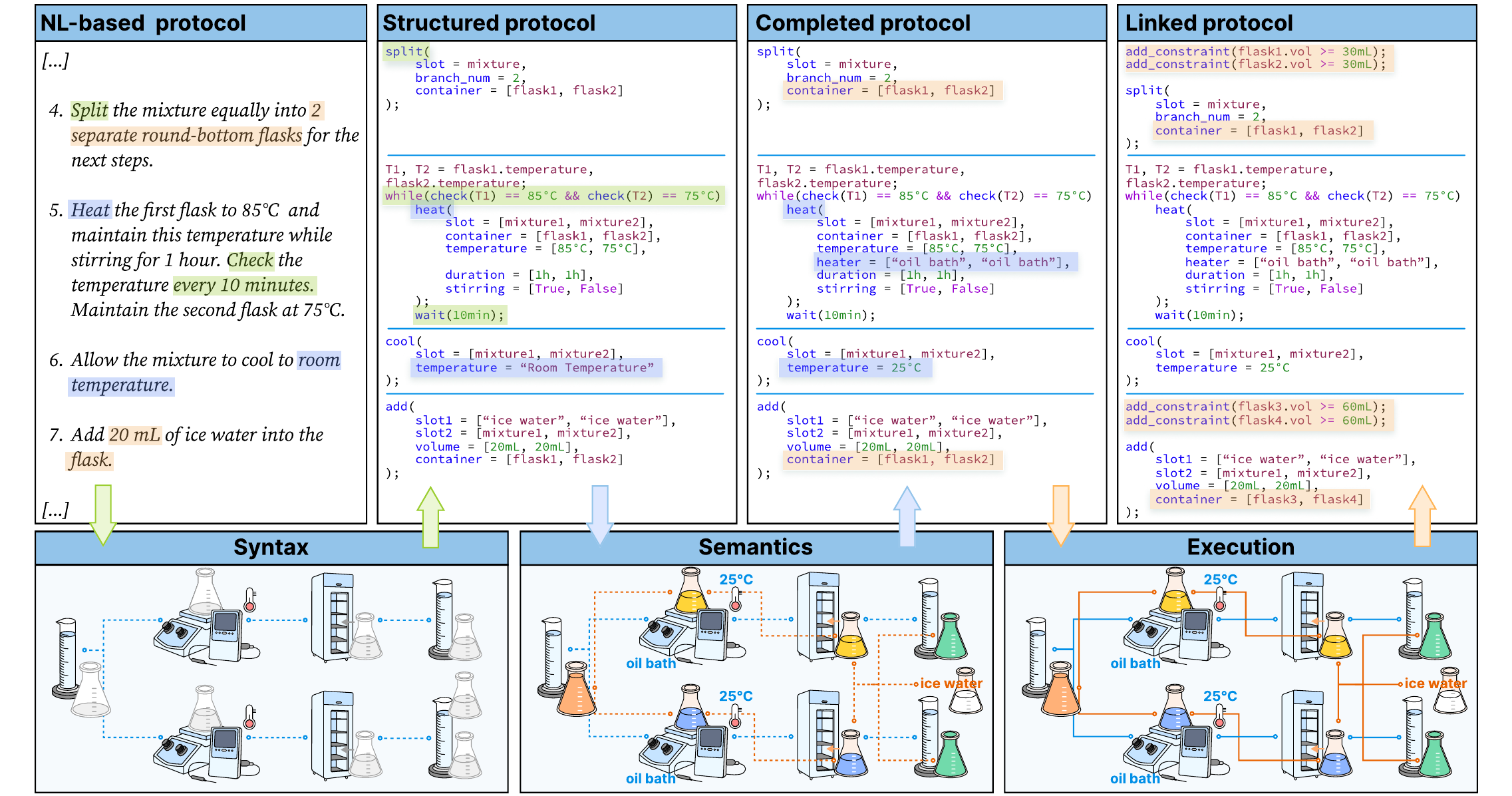}
    \caption{\textbf{Illustration of the protocol translation problem.} An \ac{nl}-based protocol is translated to a \textbf{\textcolor{fig1-syntax-green}{structured protocol}}, then to a \textbf{\textcolor{fig1-semantics-blue}{completed protocol}}, and finally to a \textbf{\textcolor{fig1-execution-orange}{linked protocol}} that is ready for self-driving laboratories along with a corresponding \ac{pdg}, after being processed through the \textbf{\textcolor{fig1-syntax-green}{syntax}}, \textbf{\textcolor{fig1-semantics-blue}{semantics}}, and \textbf{\textcolor{fig1-execution-orange}{execution}} levels. The three colors of arrows and text/ code highlights indicate the three translation steps respectively.
    }
    \vspace{-\baselineskip}
    \label{fig:overview}
\end{figure}

To bridge the aforementioned gap, the paradigm of self-driving laboratories has garnered attention, which automates experimental protocols via robotic systems, potentially revolutionizing the way experiments are conducted~\citep{bedard2018reconfigurable,steiner2019organic,mehr2020universal,rohrbach2022digitization,burger2020mobile,szymanski2023autonomous}. Despite the promising outlook, designing such labs relies largely on the translation of protocols, primarily designed for human experimenters, into machine-readable instructions. This translation process necessitates extensive collaboration between domain experts, who possess the requisite scientific knowledge; and information technology specialists, who encode this knowledge into software and hardware systems. The inherently labor-intensive nature of such translation significantly prolongs the development of self-driving laboratories. The primary challenges are rooted in the discrepancies across three critical aspects (see \cref{fig:overview}):

\paragraph{Syntax} 

Human experimenters can effortlessly \textbf{comprehend} protocols articulated in \ac{nl}, whereas automated systems frequently necessitate dedicated syntax parsers to convert these protocols into a sequence of actionable steps. Consider the protocol: \emph{``Split the mixture equally into 2 separate 50 mL round-bottom flasks for the next steps.''} This example highlights the meticulous control over experimental procedures, explicitly directing the \emph{``split''} of the mixture into precisely measured volumes --- a crucial factor for achieving uniform outcomes in subsequent reactions. It is imperative at this level to uphold a \textbf{structured} representation of the mapping of operation conditions and the control flows of operations.

\paragraph{Semantics} 

Human experimenters can \textbf{infer} implicit knowledge and context relying on the flexibility and adaptability of human understanding. In contrast, machine instructions necessitate a level of precision and rigidity that human communication does not inherently require. For instance, consider the protocol: \emph{``Stir the mixture at room temperature for 5 minutes.''}  While a human expert might inherently understand that \emph{``room temperature''} denotes a temperature range of 20-25 $^\circ C$ drawing on their prior knowledge, an automation system necessitates explicit information regarding such implicit details, which therefore need to be \textbf{completed} before execution. 

\paragraph{Execution} 

Human experimenters can \textbf{simulate} possible intermediate states and outcomes by considering the cumulative effects of a sequence of actions. For instance, given the two instructions adjacently: \emph{``Add 35 mL water to the flask``} and \emph{``Add 25 mL water to the flask''}, an experimenter can deduce that the flask's minimal capacity comes over 60 mL to prevent errors. For an automated system to perform a similar function, the actions need to be \textbf{linked} along their execution order.

Great efforts have been made on such translation tasks, among which Chemputer is representative~\citep{mehr2020universal}. This algorithm parses the \ac{nl}-based protocol into XDL, a \ac{dsl} specially designed to describe chemical synthesis reactions. The completeness and linkages are constructed with a set of manually-written constraints, with which the correctness of protocols can be further checked. This methodology has gained widespread acceptance in automated chemical synthesis, as a testament to the intensive efforts by domain and IT experts in developing XDL and the corresponding constraints. However, the application of a similar framework in other domains of experimental sciences, such as \emph{Genetics}, \emph{Medicine}, \emph{Ecology}, and \emph{Bioengineering}, would necessitate repeating these labor-intensive tasks on a case-by-case basis, thus underscoring the critical need for a more generally applicable, human-free protocol translator.

In this work, we propose a novel framework of human-free translator, designed to potentially facilitate applications across diverse experimental science domains without requiring extensive manual intervention. This framework decomposes the translation challenge into three hierarchical stages: structured on the syntax level, completed on the semantics level, and linked on the execution level, mirroring the cognitive steps undertaken by human experts in similar translation tasks. In the proposed work, the \ac{dsl}, its constraints, and linkages are generated automatically, based on protocols tailored for human experimenters, thereby eliminating the need for labor-intensive manual processes. 

Our contributions are threefold: (i) We conduct a systematic analysis of the existing discrepancies in protocol translation between human experimenters and automated systems in self-driving laboratories. From this analysis, we derive design principles that emulate human cognitive processes involved in protocol translation (\cref{sec:problem}). (ii) We devise an autonomous protocol translator through a tripartite framework that incrementally constructs \acp{pdg}, encapsulating the spatial-temporal dynamics of protocol execution across syntax, semantics, and execution levels (\cref{sec:framework}). (iii) Through both quantitative and qualitative evaluations in various experimental science domains, we demonstrate that our translator, when integrated as an auxiliary module for \acp{llm}, approaches the efficacy of skilled human experimenters and substantially surpasses the performance of purely \acp{llm}-based alternatives in protocol translation tasks (\cref{sec:result}).

\section{Protocol translation for self-driving laboratories}\label{sec:problem}

In this section, we explore the translation of protocols for human experimenters to those suitable for self-driving laboratories. We analyze the task requirements across syntax (\cref{subsec:problem-syntax}), semantics (\cref{subsec:problem-semantics}), and execution (\cref{subsec:problem-execution}) levels. We pinpoint challenges at each level for both humans and machines, delving into systematic methods for addressing these issues. Leveraging expert insights, we delineate fundamental design principles for achieving effective protocol translation (\cref{subsec:problem-design}).

\subsection{Syntax level}\label{subsec:problem-syntax}

\paragraph{Operation-condition mapping}

In \ac{nl}-based protocols, operations and their corresponding parameters such as input reagents and conditions, are entangled with each other. For example, \emph{``Dissolve 10 g of sodium chloride in 100 mL of distilled water at 80$^\circ C$''}, the entanglements of actions and conditions highlight the complexity machines face in parsing such protocols. Human experimenters can \emph{recognize} them without information loss thanks to the \emph{internalized language} for parsing \ac{nl}~\citep{chomsky1956three,chomsky2007approaching}. In contrast, protocols for machines must be represented precisely, with proper extraction of keys and values, and matching between them with appropriate data structures.

\paragraph{Operation control flows}

In \ac{nl}-based protocols, both linear and non-linear control flows are implicitly embedded in the text. While linear control flows, \ie, workflows in sequential execution order, can be straightforward, non-linear control flows such as iterative loops and parallel operations can be hard to detect because the signal and the operational domain can be separated. Consider the protocol: \emph{``Repeat the titration until the endpoint is reached, then record the volume of titrant used''}. These steps embody a non-linear control flow, challenging machines to correctly interpret the iterative process involved. Even human experts have to read the protocols carefully to understand the local and global structures to match the signals with operational domains, let alone machines. 

\begin{figure}[t]
    \centering
    \includegraphics[width=\textwidth]{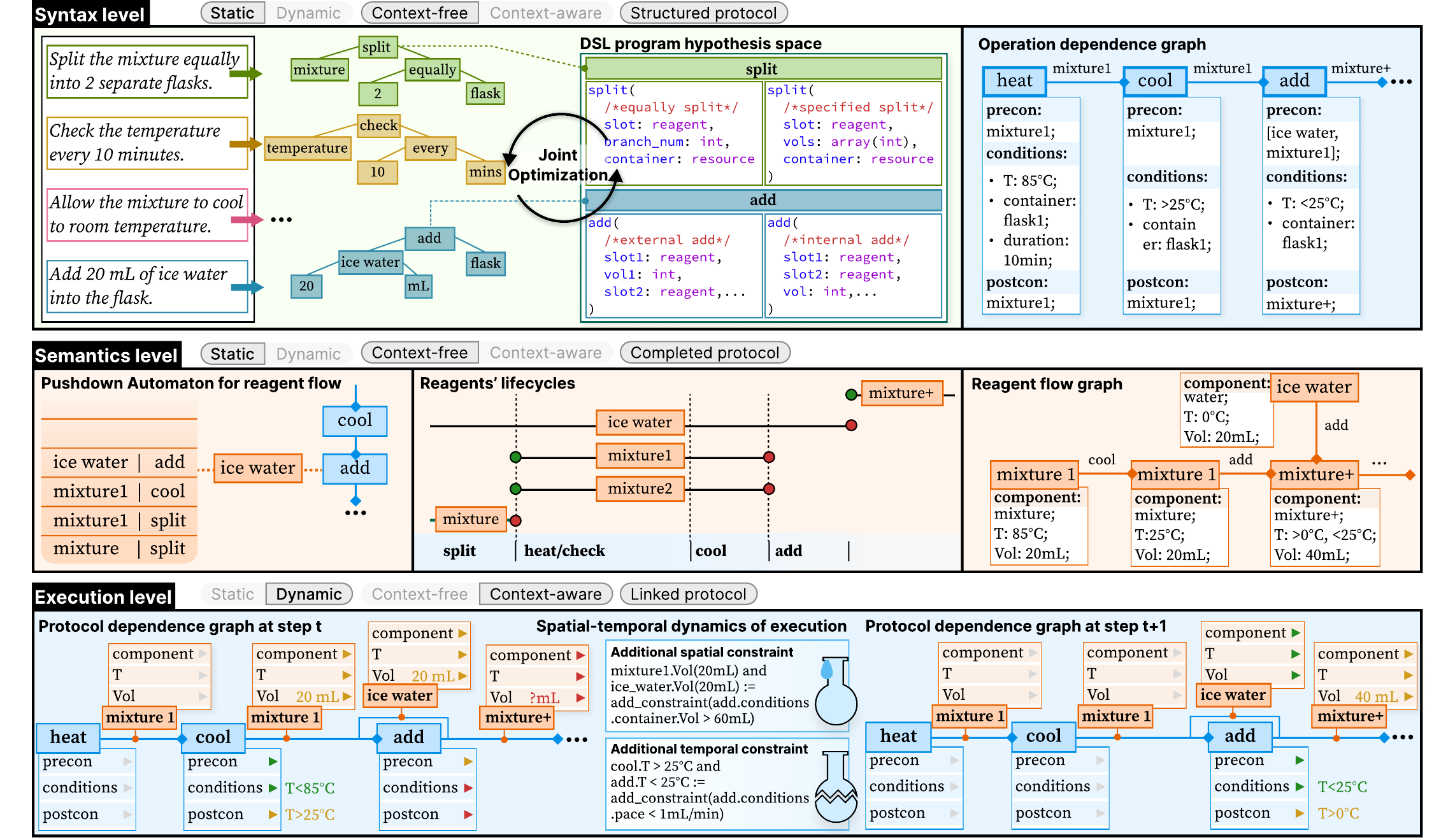}
    \caption{\textbf{The design principles and the resulting pipeline of our translator.} \textbf{(Syntax level)} Operation dependence synthesis on the syntax level, through the joint optimization of \ac{dsl} program syntax space and the parsing tree of the \ac{nl}-based protocols. This process is static and context-free. \textbf{(Semantics level)} Reagent flow analysis on the semantics level, through an automaton scheme maintaining the lifecycles of reagents and intermediate products. This process is static and context-free. \textbf{(Execution level)} Spatial-temporal dynamics analysis on the execution level, through the partial execution trace model based on the spatial-temporal dual constraint representation. This process is dynamic and context-aware.}
    \vspace{-\baselineskip}
    \label{fig:pipeline}
\end{figure}

\subsection{Semantics level}\label{subsec:problem-semantics}

\paragraph{Latent semantics of known unknowns}

Some assigned values of parameters are regarded as common sense knowledge of domain experts by default, thus the values are omitted for simplicity or referred to via a \emph{proxy name} following the domains' conventions. For example, the protocol instruction \emph{``Dry the purified product at room temperature''} relies on the experimenter's understanding of what constitutes \emph{``room temperature''}. However, machines substantially suffer from such latent semantics, implying that every value of parameters should be made explicit.

\paragraph{Latent semantics of unknown unknowns}

Sometimes, even required parameters for a specific operation are omitted from the protocols either or not intentionally, causing \emph{unknown unknowns} that one may even be not aware of the absence of such information. For instance, the protocol instruction \emph{``Centrifuge the sample after adding the enzyme''} does not specify the key controlling parameter, speed or duration, for the \emph{``centrifuge''} operation, before describing its specific value. Both human and machines require every parameter of operations to be grounded.

\subsection{Execution level}\label{subsec:problem-execution}

\paragraph{Capacity of resources}

Protocols often omit explicit specifications of resource capacities, leading to potential execution errors like exceeding a device's maximum capacity. This issue, inherent in the execution sequence, is undetectable by analyzing single operations alone. For example, the instruction \emph{``Transfer the mixture to a beaker''} requires choosing a beaker with adequate capacity, a decision based on the cumulative volume from previous steps. Humans intuitively manage this through a mental simulation of the experimental process~\citep{gallese1998mirror}. Machines, therefore, need a pre-execution verification mechanism to ensure resource capacities are not exceeded, highlighting the need for an integrated understanding of the experimental sequence.

\paragraph{Safety of operations}

In addition to managing resource capacities, another source of runtime errors stems from operations that, while semantically valid, may lead to adverse or dangerous outcomes in certain execution contexts. Such scenarios necessitate a \emph{dual-constraint} approach, where experimenters are mindful not only of the actions required \emph{what I should do} but also of potential missteps to avoid \emph{what I must not do}. For instance, the instruction \emph{``Heat the reaction mixture to 70$^\circ C$''} can be appropriate or hazardous, depending on the mixture's composition --- safe with a heat-stable catalyst, but risky with a heat-sensitive component due to potential decomposition. To navigate these complexities, human experts effectively run mental simulations, conducting \emph{"What if?"} queries and counterfactual reasoning~\citep{hoch1985counterfactual} to anticipate the consequences of their actions. Similarly, machines need a system to draw upon domain-specific knowledge and historical context to assess the safety of each operation, ensuring that all actions are contextually appropriate and safe.

\subsection{Design principles inspired by human experimenters}\label{subsec:problem-design}

Human experts' cognitive capabilities on the translation of protocols serve two key roles: understanding protocols for in-hand experiments and manually developing translators for self-driving laboratories. Inspired by these practices, we outlined design principles for our translator and assessed the strengths and weaknesses of current \acp{dsl} for \ac{nl}-based protocols, such as XDL~\citep{steiner2019organic}, ULSA~\citep{wang2022ulsa}, ORD~\citep{kearnes2021open}, Biocoder~\citep{ananthanarayanan2010biocoder}, Autoprotocol~\citep{strateos23autoprotocol}, and the family of \acp{dsl} (hereinafter called ADSL) which are automatically designed by the AutoDSL tool driven by domain corpora~\citep{shi2024autodsl}.

\paragraph{Operation dependence synthesis for the syntax level}

To precisely comprehend the complicated operation-condition mappings and non-linear control flows, machines should equip with an \emph{externalized language} in parallel with humans' \emph{internalized language}~\citep{chomsky2007approaching}. A machine-recognizable language commonly possesses a \ac{cfg} which externally defines the key-value structures on different hierarchies: (i) operation as key, reagents and conditions as values; (ii) condition as key, the corresponding parameters as values; and (iii) signal of control flow as key, the corresponding operational domains as values. If a protocol can be parsed into an \ac{ast} with the \ac{cfg}, it is verified on the syntax level~\citep{hopcroft1996introduction}, resulting in the dependency structures of the operation flow (see \cref{fig:pipeline} Top). All \acp{dsl} mentioned before are context-free languages with \acp{cfg}~\citep{fowler2010domain}, echoing this design principle.

\paragraph{Reagent flow analysis for the semantics level}

Despite the merits of \acp{dsl} based on \acp{cfg}, the context-free nature hinders verification on the semantics level, which is pivotal in protocols essentially describing procedures, where the preconditions and postconditions between temporally adjacent operations can be end-to-end connected. To be specific, although a \ac{cfg} defines a structural space with hierarchies of operations, conditions, parameters, and control flows, \ie, \emph{``There must be several parameters corresponding to a condition''}, it does not constrain the mappings under the context of domain-specific knowledge, \ie, \emph{``There are parameters controlling the Temperature, Duration, and Acidity of the condition''}. If the exact mapping between keys and values cannot be specified, the self-driving laboratories can hardly be aware of the loss of completeness, \ie, being aware of the omitted conditions given an operation or the missing value given a parameter, due to the extremely large search space over all symbols given by the corresponding \acp{dsl}~\citep{gulwani2017program}. The design choices of \acp{dsl} diverge on this level, where Autoprotocol only supports verification on the syntax level and does not possess any domain knowledge, ORD and ULSA offer the relations between operations and conditions without more fine-grained parameters, while XDL, Biocoder, and ADSL offer the find-grained key-value relation below the hierarchy of operations without more constraints about the values, \eg, suggested values of specific parameters. Hence, we require a mechanism for completing the structures of reagent flow (see \cref{fig:pipeline} Middle), while the completeness of the fine-grained parameters is guaranteed by the \acp{dsl}.

\paragraph{Spatial-temporal dynamics analysis for the execution level}

Completion on the semantics level is conducted statically because the semantics of operations are viewed individually rather than contextualized in the execution sequence. Regrettably, such effort cannot guarantee that the protocols can be executed successfully without any errors in the run time, which is unacceptable by self-driving laboratories~\citep{christensen2021automation,seifrid2022autonomous}. One way is to have domain experts write down all of the potential bad cases as constraints and use them for verification. However, run-time errors raised in the dynamic context of operations are heavily long-tail distributed. This makes it extremely hard to predict such errors from statistical \emph{hindsight}~\citep{pearl2019seven}, \ie, the set of collected post-hoc bad cases. Thus, we leverage the powerful \emph{foresight} based on simulation, which spans the full probabilistic worlds of each operation by its semantic constraints, both on the \emph{spatial} dimension, \eg, capacity of resources captured by the reagent dependency, and the \emph{temporal} dimension, \eg, safety of operations captured by the operation dependency. The simulation is conducted along the topological order of the corresponding execution flow graph. At each operation unit, both historical operations in the same protocols and similar operations in other protocols are recalled dynamically, checking and refining the dual-constraint spaces accordingly (see \cref{fig:pipeline} Bottom). Interestingly, none of the \acp{dsl} in our discussion take this feature as part of language design and only XDL employs an external compiler with hand-crafted rules for error detection. Such consideration is reasonable because in mainstream \ac{dsl} design, verification on the execution level is not guaranteed by the \acp{dsl} themselves for design simplicity and user convenience~\citep{mernik2005and}. Consequently, we require an environment to dynamically check the correctness of execution both spatially and temporally, through synthesizing operation and reagent dependencies.

\section{The framework of protocol translation}\label{sec:framework}
 
In this section, we introduce the three-stage framework for human-free protocol translation, which gradually constructs a structural representation of protocols, called the \emph{Protocol Dependence Graph} (\ac{pdg}). The \ac{pdg} makes explicit both the operation and reagent dependencies for a protocol. Operation dependence echoes the concept of program control flow, which derives the condition of sequential, branch, or loop execution of protocol operations (\cref{subsec:framework-operation}). Reagent flow provides an explicit representation of the reagent instantiate-exploit relationships implicitly in the protocol (\cref{subsec:framework-reagent}). Additionally, we simulate the protocol execution process using the \ac{pdg}, checking and refining operation sequences under the spatial and temporal dynamics of execution (\cref{subsec:framework-s-t-d}). 

\subsection{Operation dependence synthesis for the syntax level}\label{subsec:framework-operation}

The operation dependence models the topological order for executing operations in a protocol. The procedure is executed sequentially from the first operation in the protocol to the last, unless the experimenter encounters structures that change the execution flow, such as branches and loops. In practice, we extract the operation dependence by compiling the protocol to \ac{dsl} programs.

\paragraph{Input}

The compilation process is conducted based on the corresponding \ac{dsl} $\mathcal{L}=\{\mathcal{S},\Lambda\}$. The \ac{cfg}-based syntax $\mathcal{S}=(S,V,\Sigma,R)$ includes (i) the start symbol $S$; (ii) the variable set $V=\{V_{\texttt{ctrl}},V_{\texttt{op}},V_{\texttt{cond}},V_{\texttt{par}}\}$ with placeholders for control flow signals $V_{\texttt{ctrl}}$, operations $V_{\texttt{op}}$, conditions $V_{\texttt{cond}}$, and parameters $V_{\texttt{par}}$; (iii) the set of terminals $\Sigma$, which are the grounded values of parameters; (iv) the set of production rules $R$ defining the structural space between the four variable sets. The semantics $\Lambda=(T_{\texttt{ctrl}},T_{\texttt{op}},T_{\texttt{cond}},T_{\texttt{par}},T_R)$ constrains the variables and production rules, assigning the placeholders with substantial meanings. To note, according to the design choice discussed in \cref{subsec:problem-design}, the \ac{dsl} used here should be one of XDL, Biocoder, and ADSL, \ie the \acp{dsl} with the most fine-grained structural representation compared with their counterparts. 

\paragraph{Pre-processing}

Given an input protocol $\mathbf{c}$ for translation, we first parse the \ac{nl} sentences by an off-the-shelf tool and extract the actions accordingly. Then, the extracted actions are matched with the operations $o\in T_{\texttt{op}}$ of the \ac{dsl}, according to both exact match score and semantic similarity. Afterwards, we extract the arrays of entities related to the extracted action $\mathbf{e}_t\in\mathcal{E}$ by an off-the-shelf tool, where we regard the output labels to the entities and relations as \emph{pseudo-labels} because they can possibly be noisy. Please refer to \cref{subsec:supp-implementation-preprocess} for implementation details. 

\paragraph{DSL program synthesis}

Synthesizing structural representation given unstructured signal is challenging~\citep{billard2000regression,billard2006symbolic}. Specifically, the one-to-many mappings of many \ac{dsl} operations bring uncertainty into the matching. For example, the operation \texttt{add} possesses distinct patterns: two or three input slots. This further distorts the matching of reagents, conditions, and parameters because off-the-shelf tools can hardly detect the exact categories of these entities deeply rooted in domain-specific knowledge. Since the observation and hypothesis spaces are both noisy, we propose to jointly optimize the patterns of operations and the pseudo-labels. We denote the set of all possible program patterns generated by operation $o$ as $o^*=\{\texttt{p}|o\Rightarrow^*\texttt{p},T_R\Rightarrow^*\texttt{p},\texttt{p}\in T_{\texttt{ctrl}}^*\cup T_{\texttt{op}}^*\cup T_{\texttt{cond}}^*\cup T_{\texttt{par}}^*\}$. A synthesized \ac{dsl} program is defined as $\texttt{p}(\mathbf{c})=\langle\texttt{p}(o_1),\texttt{p}(o_2),\dots,\texttt{p}(o_{|p(\mathbf{c})|})\rangle$, where $\texttt{p}(o_t)$ is the program of an operation assembled with its corresponding conditions and parameters under the selected pattern in $o^*$. Let $s(\mathbf{c})=\langle\mathbf{e}_1,\mathbf{e}_2,\dots,\mathbf{e}_{|s(\mathbf{c})|}\rangle$ represent the sequence of operation-related entities, the objective of optimization can be
\begin{equation}
    \arg\min_{\texttt{p}(o_t),\mathbf{e}_t}\sum_t\sum_{o_t^*}D\big(\texttt{p}(\mathbf{c})\big\|s(\mathbf{c})\big),
\end{equation}
where $D(\cdot\|\cdot)$ is a divergence function with three indicators: (i) the selected pattern examples should be as close as possible to the text span; (ii) the selected pattern should be as similar as possible with the extracted subject-verb-object structure; and (iii) as many labeled entities as possible should be mapped to the parameter space (see \cref{fig:results}B). Though $|o^*|$ is not a large value, the whole sequence of operations can yield an exponential complexity. Hence, to make the joint optimization tractable, we separate the search of solution into two steps in the spirit of \ac{em}: (i) Expectation: sampling programs from the legal space defined by the corresponding \ac{dsl}, with a program-size-sensitive prior $|\texttt{p}(o_t)|^{-1}$; (ii) Maximization: randomly alternating symbols in $s(\mathbf{c})$ by matching them with those in $\texttt{p}(\mathbf{c})$ and greedily select the edits that decrease the objective function.

\subsection{Reagent flow analysis for the semantic level}\label{subsec:framework-reagent}

\begin{wrapfigure}{RT}{0.51\textwidth}
    \vspace{-2.5\baselineskip}
    \begin{minipage}{0.51\textwidth}
    \begin{algorithm}[H]
    \caption{Reagent flow analysis}\label{alg:pdg-construct}
        \begin{algorithmic}
        \Procedure{Transition}{$M$, $o$}
        \LineComment{\textit{Context Transition}}
        \State \textsc{Erase}($M(\Gamma)$, $\textsc{Kills}(M(\Gamma), o)$)
        \State \textsc{Append}($M(\Gamma)$, $\textsc{Defines}(o)$)
        \LineComment{\textit{State Transition}}        
        \State $M(q) \leftarrow (M(q) \,\backslash\, o) \cup \textsc{NextOps}(o)$
        \EndProcedure
        \Procedure{\textsc{Flow}}{$p(\mathbf c)$, $M$}
        \State $R = \{\,\}$ \Comment{\textit{Set of reagent dependence}}
        \State $M(q) \leftarrow \{o_1\}$ \Comment{\textit{Initial State}}
        \State $M(\Gamma) \leftarrow \langle\,\rangle$ \Comment{\textit{Initial Memory}}
        \While{$M(q) \neq \phi$}
        \State \textsc{Transition}($M$, $M(q)$)
        \EndWhile
        \EndProcedure
        \end{algorithmic}
    \end{algorithm}
    \end{minipage}
    \vspace{-1.5\baselineskip}
\end{wrapfigure}

The \ac{dsl} programs are further contextualized by associating operations with reagent flow. Reagent flow indicates the transfer of reagents among operations, reflecting how one operation impacts the subsequent ones. We define the reagent flow following the \emph{reaching definitions} schema~\citep{alfred2007compilers}, which is commonly used to capture the life cycle of a variable in compiler design. This schema determines a set of reagents reachable at each point in a protocol, and subsequently tracks the \textit{kills} and \textit{defines} of an operation, \ie, whether a reachable reagent is consumed, or a new reagent is yielded, in an operation.

\paragraph{Input}  

We denote the reagents consumed and the intermediate products yielded by each operation $o$ as $\textsc{In}(o)$ and $\textsc{Out}(o)$ respectively. The objective is to find a set of operation pairs $\{\,\langle o_i, o_j\rangle\, |\, \textsc{Out}(o_i) \cap \textsc{In}(o_j) \neq \phi \}$ such that $\textsc{Out}(o_i)$ is required as input by $\textsc{In}(o_i)$.

\paragraph{The reaching definitions schema}

We determine the availability of a reagent at each step by locating where it is defined in a protocol when execution reaches each operation. A reagent $r$ reaches an operation $o$, if there is a path from the point following $r$ to $o$, such that $r$ is not \emph{killed}, \ie, consumed, along that path. Any reagent $r$ that reaches an operation $o$ might be killed at that point, and $o$ may yield new intermediate products $r'$ that reach future operations. Notably, according to statistics on corpora of protocols~\citep{vaucher2020automated}, for about 90\% operations of a target \ac{dsl}, if $o_i\prec o_j$ are two adjacent operations, $\textsc{Out}(o_i) \cap \textsc{In}(o_j) \neq \phi$ holds. This implies that the reagent generated by preceding operation is likely to be used and then be killed instantly by the following operation. 

\paragraph{Reagent flow analysis via operation flow traversal}

We traverse the \ac{dsl} program in execution order to leverage the reagent locality revealed from statistical results, determining the reachability and life cycle of reagents. A \ac{pda} with a random access memory is adopted to record reachable reagents as operation context, defining and killing reagents at each operation point along the computation\footnote{It is worth noting that this extended \ac{pda} with random access can be shown to be in the same computation class as Turing machines~\citep{aho1972theory}, and we employ this extended \ac{pda} due to its simplicity.}. A \ac{pda} is formally defined as a 7-tuple $M=(Q, \Sigma, \Gamma, \delta, q_0, Z, F)$, where $Q \subseteq T_{\texttt{op}}$ indicates the set of states, $\Sigma = T_{\texttt{op}}$ represents the domain of inputs, $\Gamma\subseteq T_{\texttt{par}}$ denotes possible memory elements, $\delta \subseteq Q\times \Sigma \times \Gamma \to Q \times \Gamma^*$ is the transition procedure (\textsc{Transition} in \cref{alg:pdg-construct}), $q_0 = o_1$ defines the initial state, $Z=\epsilon$ is the initial memory element, and $F$ denotes the set of accepting states. The reagent dependence construction process (\textsc{Flow} in \cref{alg:pdg-construct}) traverses the \ac{dsl} program in execution order by leveraging the \textsc{NextOps} utility, which evaluates to subsequent operations. In every transition step with input, the killed reagents are removed from the memory, and the defined reagents are added to the memory. After a reagent is killed, the pair of the operations that defined it and killed it will be added to the set of reagent flow constraints. The accepting state is reached if the memory is empty at the end of execution, \ie, all reagents defined in operations are killed by other operations. We employ state-of-the-art \acp{llm} to extract reagent entities from \ac{nl}-based protocol descriptions for the two utilities \textsc{Kills} and \textsc{Defines} through instruction-following in-context learning~\citep{wei2021finetuned, brown2020language} (refer to \cref{subsec:supp-implementation-reg-flow} for details). 

\subsection{Spatial-temporal dynamics for the execution level}\label{subsec:framework-s-t-d}

While the pre-specified \ac{pdg} analysis indicates the things \emph{should} be done to follow operation and reagent flow, we still need to care about the things \emph{must not} be done by describing the activities that may be performed and the constraints prohibiting undesired execution behavior. Therefore, we introduce a constrained-based execution model to support dynamic protocol simulation, getting grounded in the theories of process modeling and execution~\citep{dourish1996freeflow, pesic2007constraint}. 

\paragraph{A constraint-based protocol execution model}

We extend the \ac{dsl} program $\texttt{p}(\mathbf{c})$  by a constraint set $C = C_{op} \cup C_{reg} \cup C_s \cup C_t$ to construct a constraint-based execution model $S = (\texttt{p}(\mathbf{c}), C)$. The execution of a program is represented by a trace $\sigma = \langle (o_1, c_1), (o_2, c_2), \ldots, (o_{|\texttt p(\mathbf c)|}, c_{|\texttt p(\mathbf c)|})\rangle$, where the order of $o_i$ reflects the temporal sequence. The execution context $c_i$ defines the spatial environment in which each operation is performed. Each constraint $c \in S(C)$ is a predicate that maps the execution trace $\sigma$ to a binary condition denoting satisfy or not. An execution trace $\sigma$ is said to satisfy the program constraint if and only if $|\sigma| = |\texttt p(\mathbf c)|$ and $c(\sigma)$ holds for all $c \in C$.

\paragraph{Leveraging partial execution trace for spatial and temporal constraints}

Explicit constraints, namely operation and reagent flow, are easy to satisfy. Unfortunately, deriving implicit constraints, \eg, the capacity of resources and safety of operations, is case-by-case for each protocol, requiring expert efforts. We propose profiling the context through execution to derive implicit spatial and temporal constraints that meet domain-specific requirements. An execution trace is defined as \textit{partially} satisfying the constraints $C$ if it follows the operation and reagent flow; that is, for any $\langle o_i, o_{j>i}\rangle$ in trace $\sigma$, there exists at least a pair of valid operation flow path and reagent flow path from $o_i$ to $o_j$.

\section{Results}\label{sec:result}

In this section, we compete our framework with human experts on the overall translation task, and assess the utility of each component of the framework by comparing with alternative approaches.

\subsection{Experimental setting}

\paragraph{Materials}

We select 75 complicated experiments with 1,166 steps in total as the testing set, from the domains of \emph{Chemical Synthesis} (235 steps in 10 experiments; 235 in 10 for simplicity; ``Synthesis'' for abbreviation), \emph{Genetics} (396 in 34), \emph{Medical and Clinical Research} (307 in 23, ``Medical''), \emph{Bioengineering} (218 in 17), and \emph{Ecology} (10 in 1). Please refer to \cref{sec:supp-test-set} for details.

\paragraph{Expert-created protocol translation}

We recruited five groups of experienced experimenters, each specializing in a different domain, with seven participants in each group. Every participating experimenter holds at least a Master's degree related to the corresponding domain, has obtained at least six years' experience in manually conducting pre-designed experiments of the domain, has acquired elementary programming skills, and has at least heard of self-driving laboratories. These human experts are asked to translate the original \ac{nl}-based protocols for their domains into those suitable for self-driving laboratories. Their outputs are subjected to \ac{dsl}-based representations and complete \acp{pdg} to evaluate machines' behaviors, which are clearly demonstrated by a running example and the examples in the \ac{dsl} documentation. Outputs from experts are carefully cross-validated and the individual divergence between them is minimized through an expert-panel-driven workshop discussion following the established workflow~\citep{reilly2023we}. Translation results of human experts and machines are serialized and are compared through ROUGE and BLEU metrics~\citep{lin2004rouge,papineni2002bleu}. Please refer to \cref{sec:supp-ethics} for ethics considerations.

\paragraph{Alternative methods}

We compare our translator with alternative methods on the first two levels, to investigate the effects of early stages on the overall translation result. On the syntax level, we compare the syntactic synthesis method (referred to as \texttt{Ours-SY}) with \texttt{ConDec-SY}~\citep{wang2023grammar}, which synthesizes \ac{dsl} programs by \acp{llm} with external \ac{dsl} grammars as constraint, and a baseline \texttt{DSL-LLM-SY} leveraging the minimal realization used in~\citet{shi2024autodsl}. On the semantics level, we compare the deductive verification method (referred to as \texttt{Ours-SE}) with \texttt{NL-RAG-LLM-SE}, which retrieves on the embedded vector database of original \ac{nl}-based protocols, and a baseline \texttt{NL-LLM-SE} implemented by pure prompt-engineering on \acp{llm}. Since the I/Oes of all stages are unified in the pipeline, we implement an overall baseline \texttt{Best-Baseline} that combines the strongest alternative methods within the evaluation of the first two stages.

\begin{figure}[t!]
    \centering
    \includegraphics[width=\textwidth]{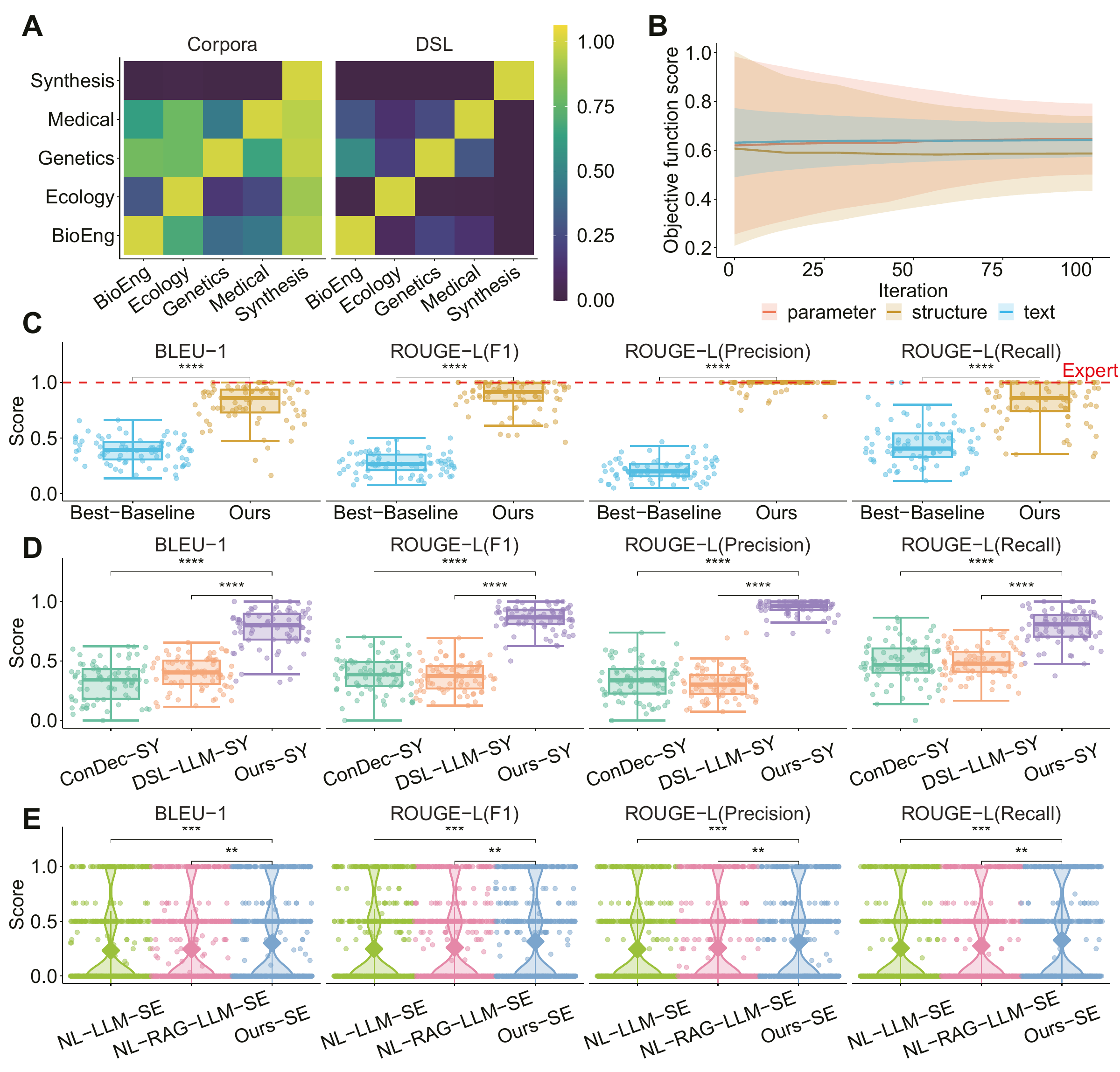}
    \caption{\textbf{Results of experiment.} \textbf{(A)} Distinctions between various domains regarding domain-specific corpora and the corresponding \acp{dsl}. \textbf{(B)} Convergence of the three indicators in the objective function for program synthesis. \textbf{(C)} Our translator significantly outperforms the best baseline and approaches human-level performance. \textbf{(D)} Our translator significantly outperforms alternative methods on the syntax level. \textbf{(E)} Our translator significantly outperforms alternative methods on the semantics level.}
    \vspace{-\baselineskip}
    \label{fig:results}
\end{figure}

\subsection{Overall assessment on expert-created protocol translation}

\paragraph{Result}

Comparing the overall output of our translator and ideal human experimenters, we find that our translator approaches the level of experts with average performance higher than $85\%$ across all indicators. Our translator significantly outperforms the alternative pipeline \texttt{Best-Baseline} on the ($t(148)=-17.71, \mu_d<0, p<.0005$; see \cref{fig:results}C).

\paragraph{Discussion}

We find that our translator demonstrates similar performance to human experts in translating protocols with complete parameters and clear descriptions (see \cref{fig:showcases}A). However, in cases where the linear description of the experimental protocol is lacking, our translator and human experts diverge. Specifically, our translator tends to translate based on the information within the sentence or between adjacent sentences, while human experts tend to consider the overall experimental process comprehensively. Though there are minor gaps, these observations suggest that our translator is approaching the level of performance of experienced human experimenters. Please refer to \cref{subsec:supp-cases-errors} for case studies on the distinctions between the behaviors of experts and machines.

\subsection{Comparison between alternative models}

\paragraph{Result}

On the syntax level, our \texttt{Ours-SY} significantly outperforms alternative approaches ($t(148)=-17.07, \mu_d<0, p<.0005$ for \texttt{ConDec-SY}; $t(148)=-15.47, \mu_d<0, p<.0005$ for \texttt{DSL-LLM-SY}; see \cref{fig:results}D). On the semantics level, our \texttt{Ours-SE} significantly outperforms alternative approaches ($t(148)=-2.52, \mu_d<0, p<.05$ for \texttt{NL-RAG-LLM-SE}; $t(148)=-3.07, \mu_d<0, p<.005$ for \texttt{NL-LLM-SE}; see \cref{fig:results}E). 

\paragraph{Discussion}

\begin{figure}[t]
    \centering
    \includegraphics[width=\textwidth]{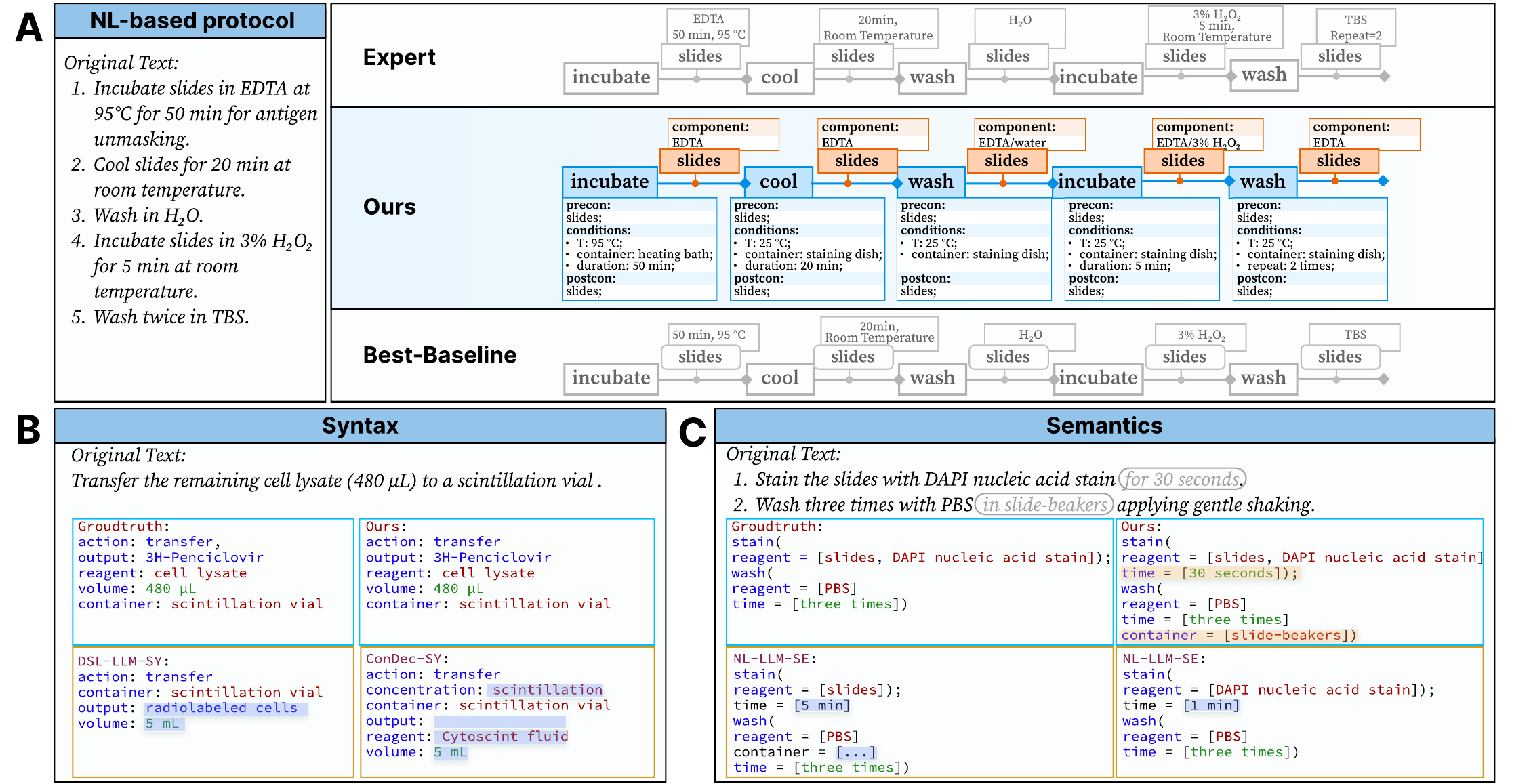}
    \caption{\textbf{Showcases of the results.} \textbf{(A)} Examples of the final \acp{pdg} generated by our translator, the alternative method, and human experts. \textbf{(B)} Examples of structured protocols output by our translator and alternative methods. \textbf{(C)} Examples of completed protocols output by our translator and alternative methods.}
    \vspace{-\baselineskip}
    \label{fig:showcases}
\end{figure}

Compared with alternative methods on the syntax level, our translator excels in ensuring the accuracy of translations across different protocols thanks to the \ac{pdg} representation, as shown in \cref{fig:showcases}B. On the semantics level, our translator performs better in translating incomplete protocols with missing information than other baselines, as shown in \cref{fig:showcases}C. We explain these merits with the properties of structural representation, which defines a representation space with high expressive power~\citep{felleisen1991expressive,lloyd2012foundations}, capturing information along the spectrum of granularity, from local details like fine-grained parameter-value relations to global-structure of control flows. By contrast, the original \ac{nl} representation and its embedding space may not possess such extent of expressive power~\citep{zhang2021neural}. Consequently, mapping the protocols into a structural latent space for query or retrieval should be more suitable than directly operating on the original space. Please refer to \cref{subsec:supp-cases-components,subsec:supp-cases-running} for details on the utilities of different approaches.

\section{General discussions}\label{sec:general-discuss}

In this work, we study the problem of translating protocols for human experimenters into those proper for self-driving laboratories. We design a human-free protocol translator under the inspiration of human experts' cognitive processes in protocol translation. Results suggest that our translator is comparable with experienced human experimenters in protocol translation, further implying its potential to serve as a plug-and-play module for fully automated scientific discovery.

\paragraph{Closing the loop of automatic scientific discovery}

The realization of automatic protocol translation closes the \emph{last-mile} of closed-loop scientific discovery. Integrating the translator with other cutting-edge techniques, we can expect such a pipeline in the future: after \ac{ai} models for scientific discovery have output insights from initial observations, there is a self-driving laboratory ready for producing and testing the designs. The creation of this laboratory is also automatic~\citep{shi2024abstract}. In the conventional creation process, we must employ a cohort of experienced domain experts to hand-craft a \ac{dsl} tailored for representing protocols of the target domain, and then hand-craft a translator with pre-defined production rules and constraints to translate the protocols into machine executables. Hence, the status quo has been changed: we can exploit the AutoDSL tool to obtain the \ac{dsl} based on a corpus of the target domain~\citep{shi2024autodsl}, and then set up our automatic translator for final execution; we can also integrate the \acp{dsl} and our translator into \ac{llm}-based ``scientist agents'' for a more reliable and flexible domain specification~\citep{boiko2023autonomous,bran2023augmenting}. The integration of these approaches facilitates the grounding of \ac{ai} for scientific discovery, which helps \ac{ai} researchers test, produce and deploy their discoveries more seamlessly. 

\paragraph{Valuing domain experts}

One might worry that the full realization of closed-loop scientific discovery would pose a severe impact on conventional experimental scientists. Indeed, \ac{ai} for scientific discovery and self-driving laboratories do not compete with conventional scientific discovery. As interdisciplinary methodologies, they still demand the supervision of scientists for tacit knowledge, creativity, and integrity~\citep{shi2024constraint}. Also, such techniques free domain experts from those time-consuming and labor-intensive low-level workloads and focus them on high-level thinking.

\section*{Acknowledgements}

This work was partially supported by the National Natural Science Foundation of China under Grants 91948302 and 52475001. Part of the authors are visiting students at Peking University during this work. In particular, Z. Bi is visiting from Huazhong University of Science and Technology and Q. Xu is visiting from University of Science and Technology of China. The authors would also like to thank Jiawen Liu for her assistance in figure drawings.

\bibliographystyle{apalike}
\bibliography{references}

\clearpage
\newpage
\appendix

\renewcommand\thefigure{A\arabic{figure}}
\setcounter{figure}{0}
\renewcommand\thetable{A\arabic{table}}
\setcounter{table}{0}
\renewcommand\theequation{A\arabic{equation}}
\setcounter{equation}{0}
\pagenumbering{arabic}
\renewcommand*{\thepage}{A\arabic{page}}
\setcounter{footnote}{0}

\section{Additional remarks}

\subsection{Rationale for the evaluation metrics}

Direct comparisons across entire sentences under BLEU and ROUGE scores would indeed pose a problem --- it may be problematic considering instructions that look similar could have very different semantics. Therefore, to circumvent this issue, we convert all results into a standardized JSON-style format for data representation, and comparisons are made between key-value pairs rather than entire sentences, effectively resolving the metric concern.

Let us consider comparing similarity between the following two instructions \emph{``... pour hot water ...''} and \emph{``... pour cold water ...''}. We transform them into the following JSON-style format:
\begin{lstlisting}[]
{
    ...
    function_name: "pour",
    reagent: "water",
    temperature: "hot",
    ...
}
...
{
    ...
    function_name: "pour",
    reagent: "water",
    temperature: "cold",
    ...
}
\end{lstlisting}

The comparison between the two sentences is then transformed into a comparison between two JSON code blocks. We calculate the similarity score cumulatively based on the similarity between the values of matched pairs of keys. For instance, for the key ``temperature'', the values ``hot'' and ``cold'' yield a low similarity score under the ROUGE, BLEU, and even the Exact Match metrics. As ``temperature'' is one of the major keys within configuration parameters, a high penalty in this dimension significantly affects the cumulative similarity score. With this fine-grained comparison metric, we can comprehensively track the distinctions and commonalities between results without losing expressivity regarding the quantities.

We also acknowledge that there are advanced evaluation metrics, especially in the recent works where \acp{llm} are leveraged as external judges and achieve considerable performance in general testing cases. Our choice of \emph{less advanced} metrics is driven by the intention to focus specifically on domain-specific knowledge, which constitutes the primary scope of this paper and may be relatively sparse in general \acp{llm}. Nonetheless, the exploration of more sophisticated evaluation metrics represents a promising avenue for future research.

\subsection{Insight behind the design of PDG}\label{subsec:pdg-insight}

We have constructed the operation dependence graph on the syntax level and the reagent flow graph on the semantics level. Indeed, the two analytical results come with a \emph{duality}. In the operation dependence graph, vertices represent operations and edges represent reagents passed between them. In contrast, the reagent flow graph uses vertices for reagent states and edges for operations causing state transitions. Interestingly, the vertices of one of the two graphs can be one-to-one mapped to the edges of another, echoing the duality. On a higher level, we say that the former provides an \emph{experimenter-centered view} while the latter offers a \emph{reagent-centered view}. These two perspectives are complementary on encoding both the information of the interventions to the environment and the status of the environment itself. Consequently, by leveraging such duality, we are able to track spatial dynamics, \eg, the variance of required resources, and temporal dynamics, \eg, the context of sequential operations, simultaneously on the \ac{pdg}. 

\subsection{Computational complexity of the framework}

Let us consider a new coming protocol with $k$ steps, with each step configured by a constant number of parameters, denoted as $c$. On the syntax level, the primary computation bottleneck arises during DSL program synthesis, where the \ac{em} Algorithm exhibits a worst-case complexity of $O(c^k)$. This is a highly conservative estimate, as mainstream optimization approaches can solve the \ac{em} much more efficiently. On the semantics level, the bottleneck occurs during reagent flow analysis, which consumes $O(k^2)$ complexity. Notably, only approximately 10\% of the steps are included in the nested loop for reagent flow construction, as about 90\% of the steps are linearly connected. On the execution level, the protocol execution model also exhibits $O(k^2)$ complexity, encompassing both forward and backward tracing. This can be optimized by replacing the full tracing strategy with a sliding window built upon the topological dependencies between steps. Although the complexities of the algorithms at these three levels are tractable, there is substantial room for improving the efficiency of the framework. Investigating methods to speed up the translation process for protocols with extremely high complexity would be a valuable area of research.

\subsection{Generality of the framework}

The general applicability of our proposed framework beyond experimental sciences can indeed be a common concern. The core value of translating \ac{nl}-based protocols into formats suitable for machine execution substantially lies in facilitating experiments in self-driving laboratories, thereby accelerating scientific discovery. Experimental protocols come with unique properties and challenges, such as the fine-grained incorporation of domain-specific knowledge, the non-trivial dependency topologies between operations, the long-horizon lifecycles of intermediate productions, and the necessity for precise execution without run-time errors. These factors shape the scope of our research problem, emphasizing the need to handle protocols with stringent terminology and formatting.

Despite the specific scope of this paper, we are open to exploring the potential for generalizing our framework to other domains with similar properties and challenges as those found in scientific experiments --- such as cooking. Imagine a self-driving kitchen that automatically prepares all ingredients and executes all procedures for cooking a meal according to \ac{nl}-based recipes. Such self-driving kitchens would also benefit significantly from translating human-oriented recipes into formats suitable for machine execution. In the following, we present a running example of such a translation, adapted from a use case of the Corel\footnote{Visit \url{https://fse.studenttheses.ub.rug.nl/25731/} for documentation.} \ac{dsl}.

The protocol after pre-processing is as follows.
\begin{lstlisting}[]
Pasta Bolognese

Yield: 2 plates

Ingredients:

- 8 [ounces] white fresh {pasta}

- 1 [floz] olive {oil}

- 1/4 [ounce] {garlic}; minced

- 4 [ounces] {onions}; chopped

- 4 [ounces] shallow fried {beef}; minced

- 1 - 1 1/2 [ounce] lean prepared {bacon}

- 1/3 [cup] red {wine}

- 150 [gram] raw {carrots}; thinly sliced

- 2/3 [ounce] concentrated {tomato puree}

- 4 [ounces] red {sweet pepper}; cut julienne

- 1 [ounce] {parmesan} cheese

Instructions:

Add the @oil@ to a large saucepan, heat to <300 F>, and saute the @onions@. 

After |2 minutes|, add the @garlic@. Keep on medium to high heat, and don't stir. 

After |2 minutes| more, add the @beef@.

Fry the @bacon@ in a separate pan, on high heat. Remove liquified fat when done.

Boil @pasta@ in a medium pan, until al dente (~|8 minutes|). Drain when done.

Once the @beef@ is done, add the @carrots@, @sweet pepper@ and @tomato puree@. 

Slowly add the @wine@ as well, to not lower the temperature. Let it simmer (but not boil) for |5-10 minutes|. 
\end{lstlisting}

Given the protocol as the input of our framework, the resulting \ac{dsl} program is as follows.
\begin{lstlisting}
add(slot = "oil", target = "large saucepan", container = plate_1, emit = mixture_1);

heat(target = mixture_1, temperature = 300F, container = plate_1, postcon = stop());

saute(target = "onions", container = plate_2, duration = 2mins);

add(slot = "garlic", target = mixture_1, container = plate_1, emit = mixture_2);

heat(target = mixture_2, temperature = 325F, container = plate_1, duration = 2mins);

add(slot = "beef", target = mixture_2, container = plate_1, emit = mixture_3);

heat(target = mixture_2, temperature = 325F, container = plate_1, postcond = check_done(target = "beef"));

fry(target = "bacon", temperature = 350F, container = pan_1, postcond = remove(target = "liquified fat"));

boil(target = "pasta", temperature = 212F, container = pan_2, duration = 8mins, postcond = drain());

add(precond = check_done(target = "beef"), slot = ["carrots", "sweet pepper", "tomato puree"], target = mixture_3, container = plate_1, emit = mixture_4);

add(slot = "wine", target = mixture_4, container = plate_1, pace = 1mL/s);

simmer(target = mixture_4, temperature = 211F, duration = 7.5mins);
\end{lstlisting}

In this example, we observe that the \ac{nl}-based recipe possesses ambiguities and omissions. Our translation framework addresses these challenges by structuring the recipe on the syntax level, completing the latent information on the semantics level, and linking the programs with necessary resources, such as the usage of plates, on the execution level. Due to the modularity of \acp{dsl}, although Corel's distribution of syntactic and semantic features differs significantly from those of \acp{dsl} used for representing experimental protocols, our translator can generalize to this new target domain through the structure of rules, namely \emph{rule-based generalization}~\citep{shi2023complexity}.

\subsection{The motivations behind this work}

In this work, we study the problem of translating experimental protocols designed for human experimenters into formats suitable for machine execution. Our primary motivation is to bridge the existing gap between machine learning algorithms in the field of \ac{ai} for science, such as molecular design, and the grounded experimental verification facilitated by self-driving laboratories. Conventional workflows for setting up self-driving laboratories and conducting physical experiments necessitate deep integration with domain experts, significantly impeding the progress of machine learning researchers in verifying and iterating their findings. Consequently, our framework aims to provide an infrastructure that enables these researchers to advance their machine learning algorithms and seamlessly validate their findings, thereby closing the loop of automatic scientific discovery.

To meet the requirements of such infrastructure, we conduct a systematic study to identify existing gaps in protocol translation between human experimenters and automatic translators in self-driving laboratories. From the study, we derive design principles that emulate human cognitive processes involved in protocol translation. Under the guidance of these design principles, we develop the three-stage framework that integrates cognitive insights from human experts with approaches from program synthesis, automaton construction, and counterfactual analysis. On the syntax level, we synthesize the operation dependence graph to transform \ac{nl}-based protocols into structured representations, thereby making explicit the operation-condition mappings and the control flows. On the semantics level, we analyze the reagent flow graph to reconstruct the complete lifecycles of intermediate products, addressing the latent, missing, or omitted properties and values. On the execution level, we contextualize both the operation dependence graph and the reagent flow graph within spatial and temporal dynamics, resulting in the protocol dependence graph. This graph conducts counterfactual reasoning to detect potential conflicts or shortages of execution resources and to identify inappropriate combinations of operations in execution sequences.

\section{Ethics statement}\label{sec:supp-ethics}

\subsection{Human participants}

The meta-evaluation included in this work has been approved by the \ac{irb} of Peking University. We have been committed to upholding the highest ethical standards in conducting this study and ensuring the protection of the rights and welfare of all participants. We paid the domain experts a wage of \$22.5/h, which is significantly higher than the standard wage. 

We have obtained informed consent from all participants, including clear and comprehensive information about the purpose of the study, the procedures involved, the risks and benefits, and the right to withdraw at any time without penalty. Participants were also assured of the confidentiality of their information. Any personal data collected (including name, age, and gender) was handled in accordance with applicable laws and regulations. 

\subsection{Corpora collection}

We carefully ensure that all experimental protocols incorporated into our corpora strictly adhere to open access policies, governed by the Creative Commons license. This approach guarantees full compliance with copyright and intellectual property laws, eliminating any potential infringement or unauthorized use of protected materials. By exclusively utilizing resources that are freely available and legally distributable, we uphold the highest standards of ethical conduct in research, fostering an environment of transparency and respect for the intellectual property rights of others. This commitment ensures that our work not only advances the frontiers of knowledge but does so in a manner that is both legally sound and ethically responsible.

\section{Implementation details}\label{sec:supp-implementation-details}

\subsection{Details of pre-processing}\label{subsec:supp-implementation-preprocess}

We employ the SpaCy Dependency Parser to analyze the syntactic structure of protocol $\mathbf{c}$, which allows for the extraction of verbs and the identification of associated objects and modifiers~\citep{honnibal-johnson-2015-improved}. After parsing, these verbs are aligned with corresponding operational actions $o\in T_{\texttt{op}}$ in the \acp{dsl} by maximizing the cosine similarity between their word2vec representations and those of the \ac{dsl} operations. Furthermore, we utilize an advanced few-shot \ac{ner} algorithm, based on large language models, to accurately identify and classify entities $\mathbf{e}_t\in\mathcal{E}$ within the text~\citep{xie2024selfimproving}. The rationale of integrating \acp{llm} with classical parsing techniques lies in leveraging the advanced natural language processing capabilities of \acp{llm} while mitigating their inherent uncertainties.

The prompt for \ac{ner} is as follows.
\begin{lstlisting}[]
Given entity label set: {label_set}.
Please name the entities in the given text. Based on the given entity label set, provide answer in the following JSON format: [{"Entity Name": "Entity Label"}]. If there is no entity in the text, return the following empty list: [].
Please note that entities have already been annotated with [], no need to extract and analyze other entities.
{cases}
Text: {query}
Answer:
\end{lstlisting}

\subsection{Details of reagent flow analysis}\label{subsec:supp-implementation-reg-flow}

We extract reagents from the natural language descriptions of protocols using two utility functions, \textsc{Kills} and \textsc{Defines}. \textsc{Kills} identifies the reagent consumed in an operation, while \textsc{Defines} identifies the reagent introduced in an operation. Due to the potential for a single chemical substance to have multiple names among other factors, it is impractical to rely solely on string matching to determine if a reagent is killed in a given operation. Instead, we employ a method based on prompt engineering with \acp{llm} for this analysis.

The following prompt is used to analyze whether the input of the current operation is the output of a previous operation:
\begin{lstlisting}[]
This instruction describes a step in an experimental process, which includes one action, multiple parameters, and one output. 
Please help analyze the output of this instruction. I will provide a list of potential outputs. You need to assist in determining which of these outputs is most suitable for this instruction. 
Note that you must choose one output from the list. Please output only a string without any explanation.

[Examples]
Instruction: {"action": "add", "reagent": ["glycoblue"], "output": ""}
Potential output list: "RNA", "mRNA"
Output: "RNA"

Instruction: {"action": "add", "concentration": ["1:10 volume 5 M NaCl"], "output": ""}
Potential output list: "a μMACS column", "solution"
Output: "solution"

Instruction: {"action": "heat", "reagent": ["limestone"], "output": ""}
Potential output list: "water", "NaCl"
Output: 

[Question]
Instruction: {Instruction}
Potential output list: {Input}
Output:
\end{lstlisting}

Additionally, this prompt is used to determine if reagents in the current memory $M(\Gamma)$ are killed by the current operation:
\begin{lstlisting}[]
This instruction describes a step in an experimental process, which includes one action, multiple parameters-including various reagents-and one output.
Please help analyze the missing reagents of this instruction. I will provide a list of potential reagents. You need to help me analyze which of these reagents might be the ones omitted from the current instruction. 
Please note how many reagent parameters are missing from the current instruction. It is possible that some reagent parameters cannot be completed with the list provided. Please output only a comma-separated list of strings without any explanation.

[Examples]
Instruction: {"action": "add", "reagent": [""], "output": ""}
Potential reagent list: "RNA", "glycoblue"
Reagents: "glycoblue"

Instruction: {"action": "add", "concentration": ["1:10 volume"], "reagent": ["", ""], "output": ""}
Potential reagent list: "μMACS", "solution", "NaCl"
Reagents: "NaCl", "μMACS"

Instruction: {"action": "use", "reagent": ["BamHI", "XhoI", ""], "device": ["PCR amplification"], "output": ""}
Potential reagent list: "agar", "food"
Reagents:

[Question]
Instruction: {Instruction}
Potential reagent list: {Memory}
Reagents:
\end{lstlisting}

In the process of synthesizing operation dependencies on the syntax level, we implement the \textsc{Defines} function. This involves pattern matching after pre-processing to accurately define reagents in each operation.

\subsection{Cost of the implementation}\label{subsec:cost}

The computational cost of our algorithm primarily arises from the expenses associated with API calls to \acp{llm}. We selected OpenAI's \texttt{gpt-3.5-turbo-0125} model for our experiments. Across 75 test protocols, we executed 1816 queries to achieve syntax-level translation, resulting in structured protocols. At the semantic level, we conducted 4062 queries for completion tasks (including translating protocols retrieved from training dataset). During these experiments, the cost model charged US\$0.50 per million tokens for inputs and US\$1.50 per million tokens for outputs. Consequently, our expenditures were approximately US\$10. Additionally, we utilized OpenAI's \texttt{text-embedding-ada-002} model to embed the training dataset and build a vector database, which incurred a cost of about US\$7.

\section{The testing set}\label{sec:supp-test-set}

\subsection{Collection}

The real experiments for the testing set are retrieved from open-sourced websites run by top-tier publishers, including Nature's Protocolexchange\footnote{\url{https://protocolexchange.researchsquare.com/}}, Cell's Star-protocols\footnote{\url{https://star-protocols.cell.com/}}, Bio-protocol\footnote{\url{https://bio-protocol.org/en}}, Wiley's Current Protocols\footnote{\url{https://currentprotocols.onlinelibrary.wiley.com/}}, and Jove\footnote{\url{https://www.jove.com/}}.

\subsection{Showcases}

\input{listings/dataset_showcases.sty}

\subsection{Instruction for human experts}

\begin{lstlisting}[]
Instruction for Human Study on Protocol Translation and Parameter Completion
[Objective]
The purpose of this study is to evaluate the accuracy and completeness of translating natural language laboratory protocols into a structured JSON representation and to assess the manual completion of missing parameters within these protocols.
[Experimental Tasks]
Participants in this study will perform two main tasks:
1. Translation of Natural Language Protocols to JSON-Structured Representation
2. Manual Parameter Completion in JSON-Structured Protocols
[Task 1: Translation of Natural Language Protocols to JSON-Structured Representation]
[Description]
Participants will be provided with a set of laboratory protocols written in natural language. The task is to translate each protocol into a JSON-structured format. This involves accurately mapping the operations, input reagents, and conditions specified in the natural language description to a precise JSON schema.
[Procedure]
1. Read the provided natural language protocol carefully.
2. Identify and extract the key elements of the protocol, including: Operations (e.g., dissolve, mix, heat)/Input reagents (e.g., sodium chloride, distilled water)/Conditions (e.g., temperature, time, concentration)
3. Construct a JSON representation that clearly reflects the structure and content of the protocol. Ensure that each element is correctly mapped to its corresponding key and value pairs.
[Example]
Extract total RNA from at least 2 x 10^6 cells using TRIZOL reagent.
{"action": "extract", "output": "total RNA", "reagent": ["TRIZOL reagent"], "volume": ["at least 2 x 10^6 cells"], "container": [""]}
[Manual Parameter Completion in JSON-Structured Protocols]
[Description]
Participants will receive a set of JSON-structured protocols with certain parameters intentionally left incomplete. The task is to manually complete these parameters based on domain knowledge and logical inference.
[Procedure]
1. Review the provided JSON-structured protocol.
2. Identify any missing or incomplete parameters.
3. Use your expertise to infer the missing information. This may include: Estimating reasonable values for missing quantities or conditions;
Ensuring consistency and coherence within the protocol.
4. Complete the JSON structure with the inferred parameters, maintaining accuracy and logical consistency.
[Example]
{"action": "apply", "output": "known DHB cluster signals", "device": ["<<<MASK>>>"]}
{"action": "apply", "output": "known DHB cluster signals", "device": ["<<<a pneumatic sprayer system>>>"]}
\end{lstlisting}

\newpage
\section{Case studies}\label{sec:supp-cases}

\subsection{Contributions of the components in our translator}\label{subsec:supp-cases-components}

We provide a series of case studies to illustrate the distinctions between the behaviors of the components within our proposed framework and those of the baselines qualitatively in \cref{tab:supp-cases-components}.

\input{tables/supp-cases-components.sty}

\newpage
\subsection{Running cases of our translator handling specific challenges}\label{subsec:supp-cases-running}

Ensuring the safety and correctness of translated protocols is an exceptionally challenging task. Several factors contribute to these challenges, including accurately mapping operations to their corresponding configuration parameters (\cref{tab:supp-cases-running-1}), precisely parsing control flows from natural language (\cref{tab:supp-cases-running-2}), completing latent semantics with domain-specific knowledge (\cref{tab:supp-cases-running-3}), inferring missing or omitted key information (\cref{tab:supp-cases-running-4}), tracking resource capacities (\cref{tab:supp-cases-running-5}), and verifying the safety of run-time execution of experiments (\cref{tab:supp-cases-running-6}). Consequently, we have made specific efforts in response to these challenges, resulting in our design of translator. Here we provide several running examples to demonstrate our translator's capability on handling the challenging factors respectively.

\input{tables/supp-cases-running.sty}

\clearpage
\newpage
\subsection{Types of errors made by our translator}\label{subsec:supp-cases-errors}

We present a detailed analysis of the errors made by our proposed automatic translator compared to human experts. We discuss the potential improvements of the translator accordingly.

\paragraph{Distinctions on the syntax level}

Difference between the translation results of our system and those of experts is subtle, with the biggest difference being in the analysis of long sentences in natural language. For human experts, it is natural and easy to analyze the parameters of events/actions or multiple actions in long sentences, while for our approach, there are sometimes problems with the correspondence between action and parameter, which need to be improved in future work. 

This series of examples in \cref{tab:supp-cases-errors-syntax-1} demonstrates the superior performance of our system at the syntax level when processing relatively short sentences.

\input{tables/supp-cases-errors-syntax-1.sty}

This series of examples in \cref{tab:supp-cases-errors-syntax-2} illustrates the challenges faced with longer sentences due to the diversity of actions and the multiple parameters.

\input{tables/supp-cases-errors-syntax-2.sty}

\newpage
\paragraph{Distinctions on the semantics level}

When supplementing known unknowns, human experts tend to rely on contextual reasoning. Since experts are not familiar with protocols from all fields, they often infer parameters based on context for protocols outside their expertise. The primary source of their errors is a lack of understanding of protocols in specific domains, which is fundamentally consistent with the approach of our system. When supplementing unknown unknowns, human experts tend to transfer their knowledge from familiar domains, such as instruments used or common parameters, to protocols in various fields, using this as a basis for parameter supplementation. Our system, however, completes parameters based on all collected protocols, which is essentially the opposite of the transfer process used by human experts.

The example presents as follows --- the completion of two types of parameters at the semantic level is included: for instance, determining the configuration parameter for an operation, where human experts rely on personal experimental experience; and inferring the required reagents for one step, where human experts use contextual reasoning. When the context is not sufficiently clear, human experts cannot infer the known unknowns within a single sentence.

\input{tables/supp-cases-errors-semantics-1.sty}

\input{tables/supp-cases-errors-semantics-2.sty}

\clearpage
\newpage
\paragraph{Distinctions on the execution level}

Human experts track capacity primarily based on prior knowledge, subsequently using context to judge the appropriateness of the equipment used. In contrast, the machine extracts the entire flow process, enabling it to calculate each step and ensure that the capacity tracking is scientifically sound and reasonable.

This series of examples in \cref{tab:supp-cases-errors-execution-1} demonstrates how our system tracks the required capacities at each step of the protocol by contextualizing the step into the spatial dimension.

\input{tables/supp-cases-errors-execution-1.sty}

This series of examples in \cref{tab:supp-cases-errors-execution-2} illustrates how our system tracks the preconditions and postconditions at each step of the protocol by contextualizing the step into the temporal dimension.

\input{tables/supp-cases-errors-execution-2.sty}

\newpage
\subsection{Properties of the pre-processing pipeline}\label{subsec:supp-cases-preprocess}

Significant differences exist between various stages of the pre-processing pipeline. We present several real-world examples to illustrate these distinctions in \cref{tab:supp-cases-preprocess}.

\input{tables/supp-cases-preprocess.sty}

\newpage
\section{Reproducibility}

The project page with supplementary files for reproducing the results of this paper will be available at \url{https://autodsl.org/procedure/papers/neurips24shi.html}.

\section{Limitations}\label{sec:supp-limitations}
As a systematic study with a proof-of-concept framework, the design and evaluation of the pipeline come with limitations, leading to further investigations: 
\begin{itemize}[noitemsep,nolistsep,topsep=0pt,leftmargin=*]
    \item We majorly exploit the approaches of empirical study to observe the behavior of \acp{dsl} and human experts for extracting design principles. Can we draw theories from information theory to rigorously prove the expression capacity of \acp{dsl} and other structural knowledge representations, to advance our design choice?
    \item We majorly consider the imperative programming \acp{dsl} as the vehicle of \acp{pdg} in this work. This raises the question of whether incorporating alternative programming paradigms, such as functional and object-oriented models, could enhance the representation of complex entities within protocols, particularly the properties of reagents.
    \item Can we extend the protocol translator to a larger set of experiments, especially those with heterogeneous hardware devices such as mobile robots?
    \item Can we find similar mechanism in other critical domains with the requirements on protocol execution, such as advanced manufacturing, and generalize our translator for such applications?
\end{itemize}
With many questions unanswered, we hope to explore more on automated protocol translation for self-driving laboratories and beyond.


\clearpage
\newpage
\section*{NeurIPS Paper Checklist}

\begin{enumerate}

\item {\bf Claims}
    \item[] Question: Do the main claims made in the abstract and introduction accurately reflect the paper's contributions and scope?
    \item[] Answer: \answerYes{} 
    \item[] Justification: In this paper, we systematically study the problem of translating protocols for human experimenters into those suitable for self-driving laboratories, in order to standardize and automate the translation process. Further, we propose the initial proof-of-concept framework that fully frees domain experts from hand-crafting protocol translators. 

\item {\bf Limitations}
    \item[] Question: Does the paper discuss the limitations of the work performed by the authors?
    \item[] Answer: \answerYes{} 
    \item[] Justification: We have discussed the potential limitations at \cref{sec:supp-limitations}.

\item {\bf Theory Assumptions and Proofs}
    \item[] Question: For each theoretical result, does the paper provide the full set of assumptions and a complete (and correct) proof?
    \item[] Answer: \answerNA{} 
    \item[] Justification: No theoretical result is included in this paper.

    \item {\bf Experimental Result Reproducibility}
    \item[] Question: Does the paper fully disclose all the information needed to reproduce the main experimental results of the paper to the extent that it affects the main claims and/or conclusions of the paper (regardless of whether the code and data are provided or not)?
    \item[] Answer: \answerYes{} 
    \item[] Justification: We have provided them with the implementation details at \cref{sec:supp-implementation-details}. We will also release our codes upon acceptance.

\item {\bf Open access to data and code}
    \item[] Question: Does the paper provide open access to the data and code, with sufficient instructions to faithfully reproduce the main experimental results, as described in supplemental material?
    \item[] Answer: \answerNo{} 
    \item[] Justification: We have provided them with the implementation details \cref{sec:supp-implementation-details}. We will also release our codes upon acceptance.

\item {\bf Experimental Setting/Details}
    \item[] Question: Does the paper specify all the training and test details (e.g., data splits, hyperparameters, how they were chosen, type of optimizer, etc.) necessary to understand the results?
    \item[] Answer: \answerYes{} 
    \item[] Justification: All of them are carefully illustrated in implementation details \cref{sec:supp-implementation-details}.

\item {\bf Experiment Statistical Significance}
    \item[] Question: Does the paper report error bars suitably and correctly defined or other appropriate information about the statistical significance of the experiments?
    \item[] Answer: \answerYes{} 
    \item[] Justification: Yes. Please refer to \cref{fig:results}.

\item {\bf Experiments Compute Resources}
    \item[] Question: For each experiment, does the paper provide sufficient information on the computer resources (type of compute workers, memory, time of execution) needed to reproduce the experiments?
    \item[] Answer: \answerYes{} 
    \item[] Justification: Please refer to \cref{subsec:cost}.
    
\item {\bf Code Of Ethics}
    \item[] Question: Does the research conducted in the paper conform, in every respect, with the NeurIPS Code of Ethics \url{https://neurips.cc/public/EthicsGuidelines}?
    \item[] Answer: \answerYes{} 
    \item[] Justification: Yes. Please refer to the main text.

\item {\bf Broader Impacts}
    \item[] Question: Does the paper discuss both potential positive societal impacts and negative societal impacts of the work performed?
    \item[] Answer: \answerYes{} 
    \item[] Justification: Please refer to the general discussions at \cref{sec:general-discuss}.
    
\item {\bf Safeguards}
    \item[] Question: Does the paper describe safeguards that have been put in place for responsible release of data or models that have a high risk for misuse (e.g., pretrained language models, image generators, or scraped datasets)?
    \item[] Answer: \answerNA{} 
    \item[] Justification: The paper poses no such risks.

\item {\bf Licenses for existing assets}
    \item[] Question: Are the creators or original owners of assets (e.g., code, data, models), used in the paper, properly credited and are the license and terms of use explicitly mentioned and properly respected?
    \item[] Answer: \answerYes{} 
    \item[] Justification: Yes. Their licenses are checked and their published works are properly cited.

\item {\bf New Assets}
    \item[] Question: Are new assets introduced in the paper well documented and is the documentation provided alongside the assets?
    \item[] Answer: \answerNA{} 
    \item[] Justification: The paper does not release new assets.

\item {\bf Crowdsourcing and Research with Human Subjects}
    \item[] Question: For crowdsourcing experiments and research with human subjects, does the paper include the full text of instructions given to participants and screenshots, if applicable, as well as details about compensation (if any)? 
    \item[] Answer: \answerYes{} 
    \item[] Justification: Please refer to \cref{sec:supp-ethics} and \cref{sec:supp-test-set}.

\item {\bf Institutional Review Board (IRB) Approvals or Equivalent for Research with Human Subjects}
    \item[] Question: Does the paper describe potential risks incurred by study participants, whether such risks were disclosed to the subjects, and whether Institutional Review Board (IRB) approvals (or an equivalent approval/review based on the requirements of your country or institution) were obtained?
    \item[] Answer: \answerYes{} 
    \item[] Justification: We have obtained an approved \ac{irb} in advance. Please refer to \cref{sec:supp-ethics}.

\end{enumerate}

\end{document}